\pdfoutput=1

\documentclass[11pt]{article}
\usepackage[english,greek]{babel}
\usepackage[LGR, T1]{fontenc}

\usepackage{acl2023}

\usepackage{times}
\usepackage{latexsym}

\usepackage[T1]{fontenc}

\usepackage[utf8]{inputenc}

\usepackage{microtype}

\usepackage{inconsolata}

\title{\selectlanguage{english}GRDD: A Dataset for Greek Dialectal NLP}
\author{\selectlanguage{english}
\selectlanguage{english}
Stergios Chatzikyriakidis \\
\selectlanguage{english}
  University of Crete \\
\selectlanguage{english}
\texttt{name.surname@uoc.gr} \\
\And \selectlanguage{english}
Chatrine Qwaider \\
\selectlanguage{english}
  Chalmers University of Technology \\
\selectlanguage{english}
\texttt{chatrine.qwaider@chalmers.se} \\ \And
\selectlanguage{english}
  Ilias Kolokousis \\
\selectlanguage{english}
  University of Leipzig \\
\selectlanguage{english}
\hskip 0.85in  \texttt{surname@studserv.uni-leipzig.de} \\\AND 
 \selectlanguage{english}
 Christina Koula\\
\selectlanguage{english}
  University of Crete \\
\selectlanguage{english}
\texttt{philp0960@philology.uoc.gr} \\ \And
  \selectlanguage{english}
 Dimitris Papadakis \\
\selectlanguage{english}
  University of Crete \\
\selectlanguage{english}
 \hskip 0.2in \texttt{ philp0961@philology.uoc.gr} \And\selectlanguage{english}
  Efthymia Sakellariou \\
\selectlanguage{english}
  University of Crete \\
\selectlanguage{english}
\hskip 0.55in\texttt{philp0962@philology.uoc.gr}}

\begin{document}
\maketitle
\selectlanguage{english}

\begin{abstract}
\selectlanguage{english}In this paper, we present a dataset for the computational study of a number of Modern Greek dialects. It consists of raw text data from four dialects of Modern Greek, Cretan, Pontic, Northern Greek and Cypriot Greek. The dataset is of considerable size, albeit imbalanced, and presents the first attempt to create large scale dialectal resources of this type for Modern Greek dialects.\footnote{The dataset can be found here: https://github.com/StergiosCha/Greek\_dialect\_corpus.}  We then use the dataset to perform dialect idefntification. We experiment with traditional ML algorithms, as well as simple DL architectures. The results show very good performance on the task, potentially revealing that the dialects in question have distinct enough characteristics allowing even simple ML models to perform well on the task. Error analysis is performed for the top performing algorithms showing that in a number of cases the errors are due to insufficient dataset cleaning. 
\end{abstract}

\section{Introduction}
Standard Modern Greek (SMG) has seen a number of computational resources being developed in the last years (see \citep{papantoniou} for an overview). However, dialects of Modern Greek have not received much attention from the perspective of NLP. One of the reasons for this situation is the scarcity of dialectal resources that can be exploited computationally to train Greek dialectal models on NLP tasks. Getting hold of dialectal data is a costly procedure. However, a number of Modern Greek dialects have an online presence, and various sources of this kind of data can be identified and extracted. For example, Cypriot Greek (CG) data can be traced in a number of online forums, social media platforms and blogspot type posts. For Pontic and Cretan Greek, data are more difficult to be found online, but a number of sources can be still identified, mainly pertaining to traditional story narrations, song lyrics, older literary texts in the dialect, as well as attempts to translate classical texts into the dialect. Similar considerations apply for the Northern varieties. 

In this paper, we try to identify the places where dialectal data can be found freely on the internet for four dialects of MG. When this is done, we extract the data, and lightly pre-process them. We end up with a quite substantial dataset, albeit an imbalanced one: 1.35m words from CG, 880k words for Cretan, 282k from Pontic, and 35k words from Northern Greek. Extra control of the data is done by extracting a random sample of 10k words per dialect and asking native speakers of the respective dialects to rate them w.r.t. validity. Lastly, we use the dataset for a dialect identification task and present the results. 
\section{Related work}
Creating datasets for the computational study Modern Greek dialects has not attracted a lot of attention and as such, attempts of this sort are rather sparse. It goes without saying that the task of Dialect Identification has not seen more work either, given this resource sparseness. There is a small dataset for the Griko \cite{anastasopoulos2018} dialect.\footnote{Griko is a Modern Greek dialect spoken in Southern Italy \cite{rohlfs1977,katsoyannou1995,chatzikyriakidis2010})}. It is parallel (Italian-Griko) and annotated  with POS tags. \citet{sababa2018} design a simple Naive Bayes Algorithm to distinguish between Standard Modern Greek (SMG) and Cypriot Greek (CG) data. As part of this project, they created a dataset of around 7k CG sentences. \citet{themistocleous2019} uses a simple DNN to distinguish between SMG and CG based on acoustic properties only (in fact, via a single sonorant). The results present an improvement over classical ML algorithms. \citet{hadjidas2015} present the Cypriot version of the Multi-CAST corpus, a dataset of annotated spoken texts. Lastly, \cite{karasimos2008greed} present an attempt to provide a database with linguistic and meta-linguistic corpora based on recorded dialectal material (505 hours of recordings).

\section{The varieties}
We collected data from 4 dialects of Modern Greek:  Cypriot Greek, Pontic Greek,  Cretan and Northern Greek, plus some data from the official Standard Modern Greek (SMG) variety. 
 A few words about these varieties are in order. 

  \subsection{Standard Modern Greek} 
Standard Modern Greek (SMG) is the official language of the
Greek and Cypriot state.
According to  \cite{mackridge:1985} (1985) has its basis in the Peloponnese Greek dialects. This can be traced back to the Greek war of Independence (1821-9) where one of the first areas to be liberated was the Peloponnese, as well as the fact that 
the Peloponnese dialects (with the exception of Tsakonian) were the dialects closer to the
written language. This newly formed variety was further enriched by elements of the dialects
of dominant Greek communities of the time, notably Greeks from Istanbul and the Ionian
islands. Other factors like compulsory education, military service,  and the the advent of radio and television “have made Greece  into a melting-pot in which speakers of various kinds of
Greek have gradually sunk their linguistic differenences'' as \citet{mackridge:1985} notes.

\subsection{Cypriot Greek}
Cypriot Greek is the Greek dialect spoken by Greek Cypriots, as well as some Turkish Cypriots. The first evidence of Cypriot Greek comes from a legal text dated in the 14th century (The Assizes).\footnote{Note that the presence of Greek in  Cyprus dates back to the 11-10th century BC (Arcadocypriot). However, Cypriot Greek is not an evolution of the Arcadocypriot dialect of Ancient Greek but rather evolved from Koin\'{e} Greek.} Despite being the native language of  most (if not all) Cypriot Greeks in both Cyprus and the diaspora, the official language of the Cypriot Greek state is SMG \cite{newton:1972,tsiplakou:2009,tsiplakou:2014}. This means that 
 modern day CG stands in a diglossic relationship to SMG \cite{tsiplakou:2009,tsiplakou:2014}.  \citet{tsiplakou:2009} further shows  that this diglossic relationship in modern day Cyprus is between SMG and an emergent pancypriot Koin\'{e}. Cypriot Greek has a number of distinct features compared to SMG. We indicatively mention the distinct weak object pronoun positioning system \cite{agouraki:2001,chatzikyriakidis:2012}, as well as the use of clefts to express wide focus constructions \cite{grohmann:2006,grohmann:2009}.

\subsection{Pontic Greek}
We use the term  Pontic Greek (PG from now on) to refer to the dialect as it is spoken today in Modern day Greece, since a  form of Pontic, Romeyka Pontic is still spoken today in some villages of Trabzon and the of area in modern day Turkey \cite{sitaridou:2012,sitaridou:2013,schreiber:20186}. The roots of PG, similarly all other Modern Greek dialects (with the exception of Tsakonika), can be traced back to Koine Greek, i.e. the form of Greek used during the Hellenistic years. However, as \citet{mackridge:1987} notes, PG came to be markedly distinct to the other Greek dialects probably due to the Seljuk invasion of the 11th century that basically split the Pontus area from the other areas of the Byzantine empire.\footnote{PG along with Cappadocian Greek comprise the Asia Minor Greek dialectal group.}  PG spoken in the Pontus region was  not a uniform dialect but included several sub-dialects. A number of classifications have been proposed by the years \cite{papadopoulos:1955,triandafyllidis:1981}. 
PG has some interesting divergences from the  linguistic structure as found in SMG. Some of them include the topicalization particle \textit{pa} \cite{sitaridou:2014}, an across the board enclitic system for weak object pronouns \cite{chatzikyriakidis2010}, and the allowance of multiple \textit{wh}-fronting among other characteristics \cite{michelioudakis:2016}. 

 \subsection{Cretan Greek}
Cretan Greek is a Greek dialect spoken in the island of Crete, also derived from  Koine Greek \cite{mackridge:1985}. According to  traditional classification, the dialect belongs to the south-western dialectal group of Greek dialects \cite{hatzidakis1905}. 
Cretan Greek has some compelling features that differentiate it among the other Greek dialects. The most characteristic one is the substitution of palato-alveolars for palatalized velars
\cite{mackridge:1985}. The so-called 'Cretan-type' fronting is a process  in which the articulation is  characterized as 'pre-palatal', i.e. mid-way between a palatal and an alveolar realization \cite{manolessou2012velar}. 

 \subsection{Kozani/Grevena Greek} 
 This is the form of Greek as it is spoken in the areas of Kozani, Grevena in the northwestern Greece.  These dialects belong to a bigger group,  i.e. that of the Northern dialects. These also include the dialects spoken in Thessaly and the Northeast Aegean, among other places. Some of the characteristics of the dialects include the use of prepositional genitives to denote possession \cite{mich:2023} and unstressed mid-vowel raising and unstressed high-vowel deletion, among other characteristics \cite{ntinas,thavoris}. 

\section{The dataset}
We focused on data that are freely available on the web. These include dialectal data from blogs, websites, as well as freely available literary texts. The first step in the collection process involved the identification and collection of urls by native speakers of the dialects or people with theoretical knowledge of the dialects, where such data can be found. We created four working groups, each focusing on finding data from the respective dialect the group was responsible for. We then developed  a number of Python scripts that allowed us to extract the dialectal text in raw format. 
We then applied a number of basic pre-processing steps, in order to clean the data: firstly, we removed empty lines and  applied a character uniformity command in order for each row to have approximately the same number of characters (56,73). Moreover, we removed special characters, duplicate lines, punctuation, and extra white spaces.  The final form of the dataset is in csv format lined up two columns, one contai. Each column ends up with the label of the respective dialect. The number of words for each dialect category, after the cleaning process, is shown in table \ref{tab:my_table}.


\begin{table}[ht]
\centering
\begin{tabular}{|c|c|}
\hline
\textbf{Dialect} & \textbf{Number of words} \\
\hline
Standard Modern Greek & 6,5m  \\
\hline
Cretan & 880k \\
\hline
Pontic & 282k \\
\hline
Northern & 35k \\
\hline
Cypriot & 1,35m  \\
\hline
\end{tabular}
\caption{The table shows the number of words for each dialect in the dataset.}
\label{tab:my_table}
\end{table}

In order to check the validity of collected data, we extracted a random sample of 10000 lines for each dialect and sent it to native speakers of that respective dialect.  These were not participants of the project. We asked them to evaluate the data quality of the sample using a scale from 1 through 10 (Table \ref{tab:evaluation_data}) using their dialectal intuitions.

\begin{table}[ht]
\centering
\begin{tabular}{|c|c|}
\hline
\textbf{Dialect} & \textbf{Avg Score} \\
\hline
Cretan & 8 \\
\hline
Pontic & 7.5\\
\hline
Northern & 9.5 \\
\hline
Cypriot & 7 \\
\hline
\end{tabular}
\caption{The table shows the average score of data evaluation by native speakers of the respective dialects.}
\label{tab:evaluation_data}
\end{table}

\section{Dialect Identification}
We used the dataset on a Dialect Identification task. We tried: a) classic ML algorithms, and b) a neural network vanilla model, i.e. a BiLSTM. We used both a) and b) on the full imbalanced dataset, but also on three balanced datasets, subsets of the original that contain 5k, 20k, and 80k lines for each dialect respectively. 

\subsection{Classical ML algorithms}
We ran a number of ML algorithms using a simple word-gram language model (both 1-gram and 3-gram). We see that accuracy for the full dataset is quite high, which seems to  that the models are able to correctly identify distinctive features for the dialects and learn to classify them correctly. The results are reported below:

\begin{table}[h!]
\centering
\begin{tabular}{|c|c|}
\hline
\textbf{ML Algorithm} & \textbf{Score} \\
\hline
Ridge classifier & 0.91 \\
\hline
Naive Bayes & 0.92 \\
\hline
SVM & 0.91 \\
\hline
\end{tabular}
\caption{Top-three performing algorithms on the DI task}
\label{table:1}
\end{table}

In order to investigate the effect of data imbalance in the classification we created three versions of the original dataset with 5k, 20k and 80k lines respectively. We present the results of the top three performing algorithms for each dataset below:

\begin{table}[ht]
\centering
\begin{tabular}{|c|c|c|c|}
\hline
\textbf{Dataset/Model} & \textbf{Ridge} & \textbf{NB} & \textbf{SVC} \\
\hline
\textbf{5k} & \textbf{0.76} & \textbf{0.76} & 0.75 \\
\hline
\textbf{20k} & \textbf{0.82} & \textbf{0.81 }& 0.82 \\
\hline
\textbf{80k} & 0.89 & \textbf{0.90} & 0.88 \\
\hline
\end{tabular}
\caption{Top three performing algorithms on three balanced datasets}
\label{tab:ml_results}
\end{table}

We see a steady increase in accuracy for all algorithms as we add more data. With 80k lines per each dialect we get an accuracy of 0.90 for the NB model. This is getting close to the performance on the original imbalanced dataset (NB performs 0.94 accuracy on the full dataset). 

\subsection{BiLSTM}
Following the relative success of simple ML algorithms in the task, we experimented with a more elaborate DL model, in particular a vanilla BiLSTM model, with an 128 embedding layer, two BiLSTM layers (64 and 32 layers respectively), a densely connected network with 64 units, relu as the activation function and, lastly, a final layer with softmax activation. We  run the model using only two epochs. The results present a very high accuracy of 97\%  and low loss (approximately 10\%). In order to check for model overfitting, we performed k-fold validation using the same number of epochs. The results show a decrease in  average accuracy as well as an increase in average loss. However, the results seem to indicate that the model is generalizing well to the training data, even though some overfitting seems to be at play. The results are shown in:

\begin{table}[h]
\centering
\begin{tabular}{|c|c|c|}
\hline
\textbf{} & \textbf{Accuracy} & \textbf{Loss} \\
\hline
\textbf{BiLSTM} & 0.97 & 0.1 \\
\hline
\textbf{k-fold validation} & 0.935 & 0.17 \\
\hline
\end{tabular}
\caption{Comparison of BiLSTM and k-fold}
\label{tab:comparison}
\end{table}

We ran the same model on the smaller balanced datasets and performed k-fold validation on those too:

\begin{table}[h]
\centering
\begin{tabular}{|c|c|c|}
\hline
\textbf{} & \textbf{Accuracy} & \textbf{Loss} \\
\hline
\textbf{BiLSTM\_5000} & 0.89 & 0.5 \\
\hline
\textbf{k-fold\_5000} & 0.9 &  0.32\\
\hline
\textbf{BiLSTM\_20k} & 0.928 & 0.21 \\
\hline
\textbf{k-fold\_20k} & 0.922 & 0.2 \\
\hline
\textbf{BiLSTM\_80k} & 0.94 & 0.15 \\
\hline
\textbf{k-fold\_80k} & 0.936 & 0.18 \\
\hline
\end{tabular}
\caption{Top three performing algorithms for each balanced dataset}
\label{tab:comparison}
\end{table}

\section{Error analysis and limitations}
We took a random sample of 20 miss-classified examples for each dialect and for each ML model. The results are quite interesting. In algorithms where performance is quite low, most of the miss-classified examples are genuine cases of miss-classification, where the algorithm has not managed to find enough distinctive features to distinguish the dialects. Consider the following example, a missclassified example from the SGD algorithm:

\begin{description}
    \item[Text:]
    \begin{otherlanguage*}{greek}Αυτοί πόρχουν\end{otherlanguage*}\begin{otherlanguage*}{english}d\end{otherlanguage*}\begin{otherlanguage*}{greek}αν για καμνιά κούπα πιρνούσαν μ\end{otherlanguage*}\begin{otherlanguage*}{english}b\end{otherlanguage*}\begin{otherlanguage*}{greek}ρουστά τ τουν\end{otherlanguage*} \begin{otherlanguage*}{english}g\end{otherlanguage*}\begin{otherlanguage*}{greek}αλημέρζαν\end{otherlanguage*} \\\begin{otherlanguage*}{english}
    \textbf{Correct Dialect:} northern \\
    \textbf{Predicted Dialect:} cypriot
    \end{otherlanguage*} 
\end{description}
In the above, the classifier clearly cannot distinguish between northern varieties and cypriot, even though the example contains  typical features of northern varieties that do not appear in cypriot (front vowel raising and front vowel loss). On the other hand, consider the following:
\begin{description}
    \item[Text:]
    \begin{otherlanguage*}{greek}ότι όλα έχουν τη θέση τους και την αξία
\end{otherlanguage*} \\\begin{otherlanguage*}{english}
    \textbf{Correct Dialect:} Cypriot \\
    \textbf{Predicted Dialect:} Greek
    \end{otherlanguage*} 
\end{description}

This example represents a miss-classification of a different kind. The mistake here is due to incomplete cleaning of the dataset. The example in question is clearly in Modern Greek that  ended up in the Cypriot dataset. For example it could be part of a response to a blog post written in  Cypriot (the post) in Standard Greek. In effect, the classifier is right here, pointing to incomplete cleaning in this case. In future work, we will use this type of error analysis based on the DI task to further clean the dataset and enhance its validity and reliability.  

A further limitation of our work concerns the nature and type of data found in the dataset, in effect also connecting to the previous issue of data cleaning. We have created a dataset that is coarse-grained, in the sense that different varieties of the same dialect, different genres\textbackslash registers\textbackslash varieties in\textbackslash of the dialect, have  collapsed in one category. For example, in the case of Cypriot Greek, most of the data are collected from blog posts, which correspond to a specific text type, whereas in Cretan, a lot of the data come from translations of Ancient Greek tragedies and comedies into Cretan, and musical declamations (mantinades). However, despite this obvious limitation, we believe that the dataset can be used as a first basis to create more fine-grained datasets, where these  considerations are taken into account. For the moment, what we want to offer to the community is a dialectal dataset that can be used/extended/modified to be used for dialectal NLP tasks.  
\section{Conclusion}
We have presented a dataset of Greek dialectal data. In particular, we collected freely available data from Pontic, Cretan, Northern varieties and Cypriot, we performed some slight preprocessing and compiled the first dataset of this size that can be used for the computational study of Greek dialects. We performed a simple Dialect Identification task on the dataset. The results point out to the dialects having enough distinctive characteristics that can be traced even by very simple ML algorithms and eventually perform well on the task. A simple BiLSTM performs the best beating traditional ML agorithms even with small amounts of data. The error analysis suggests that a good portion of error lies in insufficient data cleaning. We will further work on the cleaning based on the error analysis results. We aspire this dataset to be the basis for further work and tasks on Greek dialectal NLP. 
\section*{Acknowledgements}
The Special Account for Research Funding of the Technical University of Crete is thanked for funding part of this research (grant number: 11218). Stergios Chatzikyriakidis also gratefully acknowledges funding from  the European Union under Horizon Europe (TALOS-AI4SSH 101087269). 
We also want to thank the following people for their help with this project: a) Christodoulos Christodoulou for kindly giving us access to his corpus of Kozani Greek, b) Michalis Sfakianakis for kindly giving us access to his translations of numerous ancient Greek tragedies and comedies into Cretan Greek, and c) postgraduate students Irini Amaniki, Irini Giannikouri, Efrosini Skoulataki, Erofili Psaltaki, Valentini Mamatzaki, Vassiliki Katsouli and Hara Soupiona for their help in the initial stages of identifying dialectal data on the net. 
\bibliography{ranlp2023,custom}
\bibliographystyle{acl_natbib}

\end{document}